\documentclass[10pt,twocolumn,letterpaper]{article}

\usepackage[pagenumbers]{cvpr} 

\usepackage{cuted}

\definecolor{cvprblue}{rgb}{0.21,0.49,0.74}
\usepackage[pagebackref,breaklinks,colorlinks,allcolors=cvprblue]{hyperref}

\def\paperID{11892} 
\def\confName{CVPR}
\def\confYear{2026}

\newcommand{\alg}{FastAvatar}

\title{\alg{}:\\ Instant 3D Gaussian Splatting for Faces from Single Unconstrained Poses}

\author{Hao Liang\\
Rice University\\
Houston, TX, USA\\
{\tt\small hl106@rice.edu}
\and
Zhixuan Ge\\
Rice University\\
Houston, TX, USA\\
{\tt\small zg33@rice.edu}
\and
Ashish Tiwari\\
Rice University\\
Houston, TX, USA\\
{\tt\small ashish.tiwari@iitgn.ac.in}
\and
Soumendu Majee\\
Samsung Research America\\
Dallas, TX, USA\\
{\tt\small s.majee@samsung.com}
\and
Dilshan Godaliyadda\\
Samsung Research America\\
Dallas, TX, USA\\
{\tt\small dilshan.g@samsung.com}
\and
Ashok Veeraraghavan\\
Rice University\\
Houston, TX, USA\\
{\tt\small vashok@rice.edu}
\and
Guha Balakrishnan\\
Rice University\\
Houston, TX, USA\\
{\tt\small guha@rice.edu}}

\begin{document}
\maketitle
\begin{strip}
    \\[-1.3cm]
    \centering
    \includegraphics[width=\textwidth, trim=20 0 10 0, clip]{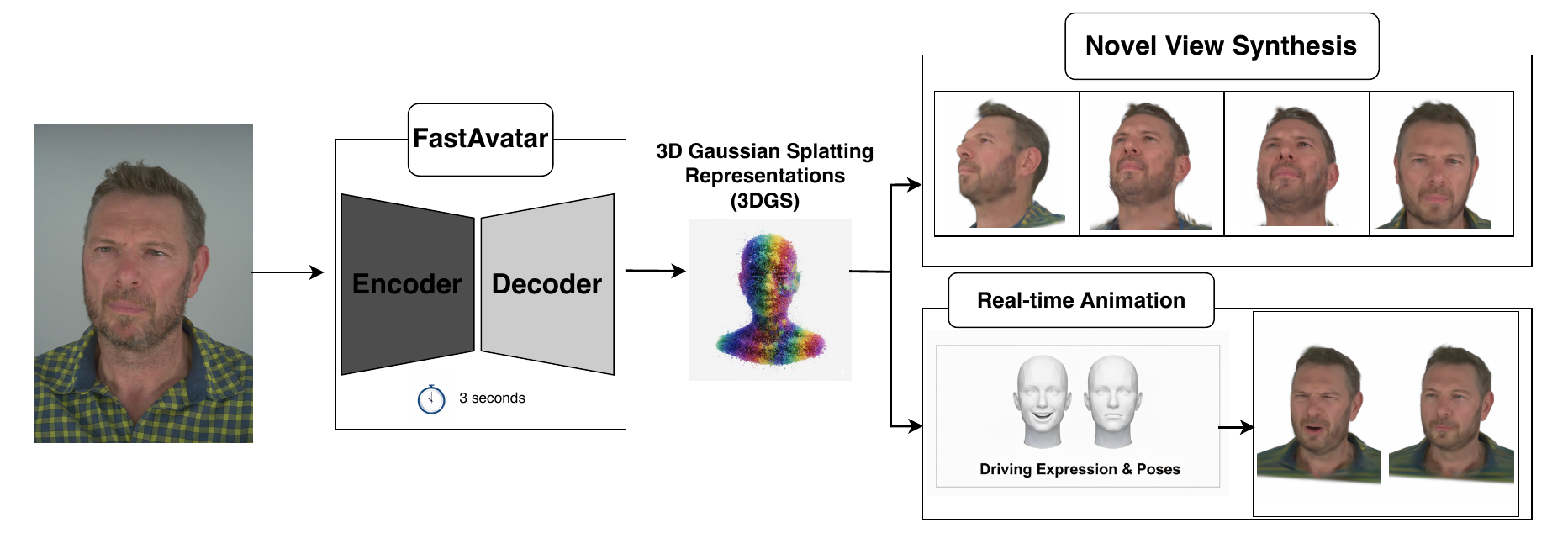}
   \captionof{figure}{\textbf{\alg{} produces high-quality 3D face avatars and animations from a single input image.} Given an arbitrary-pose face image, \alg{} reconstructs a complete 3D Gaussian Splatting (3DGS) representation and refines it using a geometry–appearance optimization routine requiring only $\sim$3 seconds on a single NVIDIA A100 GPU. Once reconstructed, the avatar supports photorealistic novel-view synthesis and smooth expression animation driven by FLAME-guided pose and expression controls, while preserving identity and rendering quality across all viewpoints.}
    \label{fig:teaser}
\end{strip}

\begin{abstract}
We present \alg{}, a fast and robust algorithm for single-image 3D face reconstruction using 3D Gaussian Splatting (3DGS). 
Given a single input image from an arbitrary pose, \alg{} recovers a high-quality, full-head 3DGS avatar in approximately $3$ seconds on a single NVIDIA A100 GPU. We use a two-stage design: a feed-forward encoder–decoder predicts coarse face \emph{geometry} by regressing Gaussian structure from a pose-invariant identity embedding, and a lightweight test-time refinement stage then optimizes the \emph{appearance} parameters for photorealistic rendering. 
This hybrid strategy combines the speed and stability of direct prediction with the accuracy of optimization, enabling strong identity preservation even under extreme input poses. \alg{} achieves state-of-the-art reconstruction quality ($24.01$ dB PSNR, $0.91$ SSIM) while running over $600\times$ faster than existing per-subject optimization methods (e.g., FlashAvatar, GaussianAvatars, GASP). 
Once reconstructed, our avatars support photorealistic novel-view synthesis and FLAME-guided expression animation, enabling controllable reenactment from a single image. By jointly offering high fidelity, robustness to pose, and rapid reconstruction, \alg{} significantly broadens the applicability of 3DGS-based facial avatars.
\end{abstract}    
\section{Introduction}
Creating 3D face models from images is a long-standing problem in computer vision and graphics, and is of significant current interest in digital avatar applications such as virtual reality, gaming, and content creation. 3D face model frameworks that enable fast and high-resolution novel view rendering performance from one or few input views are needed to support these applications. Classical parametric face models based on simple statistical approaches~\cite{blanz2003face, blanz2023morphable,FLAME:SiggraphAsia2017} offer real-time fitting and rendering speed, but are limited in their expressive power. In contrast, recent approaches based on the Neural Radiance Field (NeRF)~\cite{mildenhall2021nerf} and 3D Gaussian Splatting (3DGS)~\cite{kerbl20233d} neural rendering frameworks offer state-of-the-art expressive power, with 3DGS even enabling real-time rendering speed. However, these algorithms have critical limitations: they require expensive per-subject optimizations from multi-view captures with fitting times ranging from minutes to hours, restricting their  practical deployment in many consumer applications.

In this study, we aim to move beyond purely optimization-based avatar construction and propose a fast, robust framework that reconstructs a high-quality 3DGS face model from a single input image. Purely feed-forward face generation approaches, typically based on GANs~\cite{goodfellow2014generative} or diffusion models~\cite{ho2020denoising}, are often constrained to (near-)frontal viewpoints due to data bias and struggle to generalize under large pose variation, frequently exhibiting artifacts or identity drift when viewed from novel angles~\cite{gerogiannis2025arc2avatar,chan2022efficient}. 
At the same time, fully optimization-driven 3DGS reconstruction methods deliver high fidelity but require minutes of per-subject fitting and are unsuitable for interactive user applications. 
This motivates the need for a single-view reconstruction framework that combines the speed and stability of direct prediction with the fidelity of optimization, while remaining robust to large input pose variation and maintaining faithful identity preservation.

To address these challenges, we present \alg{}, a fast framework for 3DGS face reconstruction from a single image with arbitrary pose. \alg{} is driven by two key insights. First, inspired by classical morphable face models~\cite{blanz2003face}, we construct a 3DGS ``template'' face representation by averaging the Gaussian parameters across a set of subject-specific 3DGS models. Unlike prior work that initializes Gaussians randomly or requires per-subject optimization~\cite{saunders2025gasp,qian2024gaussianavatars,Yan_2025_WACV,liu2024dynamic}, this template provides a geometric prior that enables fast and stable reconstructions. At test time, \alg{} represents a new identity by predicting residuals to this template using an encoder–decoder network.

Second, to achieve identity preservation even under extreme viewpoints, we constrain the encoder to map all views of the same individual to a shared latent vector using pre-trained face recognition features~\cite{deng2018arcface} and contrastive learning. 
The decoder then maps this identity embedding to Gaussian parameter residuals that refine the template into a subject-specific 3D representation. This encoder–decoder architecture effectively learns a strong geometry prior over human faces: all identities are reconstructed as deformations of a shared canonical 3DGS template. 
As a result, \alg{} produces stable, 3D-consistent geometry even from extreme input poses, avoiding the view-dependent artifacts or identity drift commonly observed in unconstrained image-to-3D prediction networks. This feed-forward prediction produces the coarse \emph{geometry}, which we subsequently refine through a lightweight optimization stage to recover high-quality \emph{appearance} while preserving identity. 

We evaluated \alg{} with several quantitative and qualitative experiments using the Nersemble test dataset~\cite{kirschstein2023nersemble} and compared against recent 3DGS-based and diffusion-based avatar methods. Given three distinct input poses (frontal and two extreme views), we measured reconstruction quality across 15 novel viewpoints and observe that \alg{} achieves 24 dB PSNR in approximately $3$ seconds, outperforming feed-forward baselines such as GAGAvatar~\cite{chu2024generalizable} (16 dB), LAM~\cite{he2025lam}(14 dB), and running $600\times$ faster than purely optimization-driven approaches such as GaussianAvatars~\cite{qian2024gaussianavatars}, FlashAvatar~\cite{xiang2024flashavatar}, DiffusionRig~\cite{ding2023diffusionrig}, and Arc2Avatar~\cite{gerogiannis2025arc2avatar}. \alg{} maintains stable identity and reconstruction quality across large input poses, from frontal to extreme profiles, whereas existing feed-forward or template-free approaches tend to degrade or drift. We also provide qualitative examples using \alg{} to perform FLAME-guided expression animation from a single image. By combining high fidelity, robustness to pose, and rapid reconstruction, \alg{} broadens the applicability of 3D Gaussian Splatting to practical and interactive avatar creation.
\section{Related Work}
Recovering a high-quality 3D face from images is a long-standing problem in computer vision and graphics. 
The seminal 3D Morphable Model (3DMM)~\cite{blanz2003face,smith2020morphable,blanz2023morphable,li2020learning} represents facial geometry and appearance using statistical mesh-based bases learned from many subjects. 
FLAME~\cite{FLAME:SiggraphAsia2017} extends these models to capture expression variations and articulated jaw and neck motion using thousands of 3D head scans~\cite{feng2021learning,danvevcek2022emoca}. 
Both 3DMM- and FLAME-based methods estimate parameters via inverse rendering or neural regression, but their low-rank bases limit expressiveness for high-frequency facial details such as subtle skin variation and hair patterns.

In the past five years, the field of neural rendering has opened new capabilities in 3D face reconstruction. Neural Radiance Fields (NeRF)~\cite{mildenhall2021nerf} and their dynamic variants~\cite{pumarola2021d,park2021hypernerf,li2021neural,pumarola2021d,fang2022fast} enable photorealistic view synthesis by representing scenes as continuous volumetric fields. NeRFs have also been developed for face reconstruction 
~\cite{tretschk2021non,park2021nerfies,zhuang2022mofanerf,hong2022headnerf,gafni2021dynamic,buehler2024cafca,trevithick2023real}. However, NeRFs are typically slow to train and render, often requiring hours to fit a new scene. 3D Gaussian Splatting (3DGS) is a compelling alternative to NeRF that explicitly represents scenes as a set of anisotropic Gaussian primitives with learned parameters~\cite{kerbl20233d}. 3DGS supports high-quality rendering at real-time frame rates and has been adapted to face and head modeling, achieving impressive visual quality~\cite{xu20243d,dhamo2024headgas,wei2025graphavatar,ma20243d}. Recent studies have started to combine parametric meshes with 3DGS to gain the advantages of dense correspondence and identity-expression disentanglement from meshes, along with efficient, high-quality rendering from 3DGS~\cite{qian2024gaussianavatars,shao2024splattingavatar,xiang2024flashavatar,kocabas2024hugs,radford2021learning,xu2024gaussian,qian20243dgs,saunders2025gasp,Yan_2025_WACV,hu2024gaussianavatar,wang2025mega,giebenhain2024npga}. These methods typically attach Gaussians to vertices (or surface patches) of a fitted parametric head mesh and optimize per-Gaussian parameters (means, scales, appearances) to the observed image(s). In addition, while most of these methods require multi-view capture, several have been extended to the far more under-constrained single-view setting~\cite{ki2024learning,li2023generalizable,ma2023otavatar,tran2024voodoo,yang2020facescape,zielonka2022towards,li2024talkinggaussian,he2025lam}. Unlike prior mesh-anchored 3DGS methods that initialize templates randomly or jointly optimize them with the decoder (entangling with the latent space), we fix the template by averaging subject-specific optimized 3DGS models and only learn offsets—yielding smaller, better-conditioned residuals, removing the $(K \times P)$ parameter block, and improving stability and generalization. While these hybrid 3DGS approaches can produce detailed reconstructions and can fit scenes far faster than NeRFs (often nearly $10\times$ faster), their fitting times are still far from real-time due to their reliance on per-face iterative optimization. 

In contrast to iterative optimization approaches, feed-forward methods directly map input images to outputs --- either 3D models or novel views --- resulting in real-time fitting speed. Several feed-forward methods directly predict parameters to NeRF or 3DGS models given one or more input images~\cite{zheng2023pointavatar,li2023generalizable,li2023one,chu2024generalizable,ma2024cvthead,yang2024learning,chu2024gpavatar,ye2024real3d,Wang_2025_CVPR}. These approaches work well on frontal or near-frontal inputs with limited pose variation, but often exhibit pose-dependent inconsistencies; reconstructions from non-frontal inputs can differ noticeably in identity and geometry compared to frontal inputs. Another family of feed-forward methods use generative models such as diffusion models \cite{paraperas2024arc2face,gerogiannis2025arc2avatar,shiohara2024face2diffusion,chen2024dreamidentity,ma2024subject,shi2024instantbooth,wei2023elite,xiao2025fastcomposer,taubner2025cap4d} or GANs \cite{chan2022efficient,sun2023next3d,zhao2024invertavatar,deng2024portrait4d,karras2019style,karras2020analyzing,an2023panohead,gecer2020synthesizing,lattas2023fitme} to directly infer novel views by exploiting data-driven priors. Generative models now offer photorealistic synthesis quality, but do not typically construct explicit 3D geometry, and thus often produce distortions, drifting facial structures, and identity hallucinations with pose changes.

\alg{}, the proposed algorithm in this study, integrates several ideas developed in these existing lines of work to perform high-quality, real-time novel view synthesis given only a single input face image. \alg{} uses a feed-forward architecture to predict deformations to a template 3DGS model, taking inspiration from morphable models. The architecture also maps the input image into a pose-invariant embedding, enabling better identity-pose disentanglement given an input image from any arbitrary pose. 




\begin{figure*}[!t]
   \includegraphics[width=\textwidth]{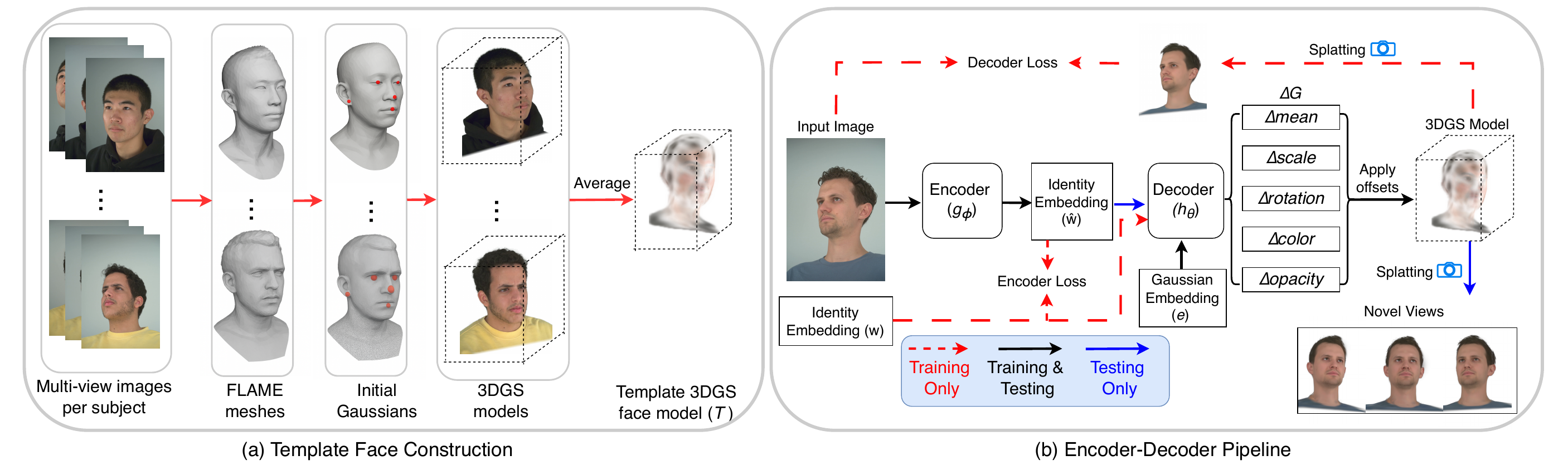}
   \caption{\textbf{FastAvatar framework.}
    \textbf{(a) Template 3DGS face model construction.} \alg{} constructs a template 3DGS face model $\mathcal{T}$ by averaging parameters of Gaussians across 3DGS models fit on a training set of subjects.
    \textbf{(b) Encoder-Decoder Pipeline.} \alg{} uses an encoder-decoder architecture to map an input image to parameter offsets of the template 3DGS model constructed in (a). We train the decoder to predict parameter offsets for each Gaussian conditioned on subject-specific and Gaussian-specific embedding vectors. We train the encoder to map multi-pose images of the same identity to the same subject-specific embedding. At inference time, \alg{} passes an image into the encoder to generate a subject-specific embedding, and decodes this embedding to obtain Gaussian-specific parameter offsets, that, combined with template $T$, yields a full 3DGS avatar in real time ($\leq 3 $ seconds) with refinement.}
   \label{fig:pipeline}
\end{figure*}
\section{Method}
Given a single face image $I$, our goal is to reconstruct a complete 3DGS model that enables high-quality novel-view synthesis and animation. The main challenge is to infer thousands of Gaussian parameters from a single 2D observation under extreme pose variation. \alg{} addresses this using a canonical 3DGS template and pose-invariant residual prediction. We first build a template by averaging Gaussian parameters across subject-specific 3DGS models, with all Gaussians placed at consistent FLAME-tracked locations. This provides a strong geometric prior and ensures consistent correspondence across subjects.

At test time, an encoder maps the input image to a pose-independent identity embedding, and a decoder predicts per-Gaussian residual parameters that deform the template into a subject-specific model. 
For non-position attributes (opacity, scale, rotation, SH appearance), the decoder predicts residuals from the template values. For positions, it predicts offsets from the canonical FLAME-tracked locations to maintain stable geometry under large viewpoint changes. This feed-forward prediction yields a coarse 3DGS geometry, which is further refined by a lightweight appearance optimization stage for accurate rendering. 

\subsection{Preliminaries: 3D Gaussian Splatting (3DGS)}
A 3D Gaussian Splatting (3DGS) model~\cite{kerbl20233d} represents a scene using $K$ anisotropic Gaussians 
$\mathcal{M}=\{G_k\}_{k=1}^K$, each defined by geometric and appearance parameters: center $\mu_k\in\mathbb{R}^3$, 
opacity $\alpha_k\!\in[0,1]$, spherical harmonic (SH) color coefficients $c_k$, and a covariance matrix 
$\Sigma_k\!=\!R_k S_k^2 R_k^T$ factored into a rotation $R_k$ (quaternion $q_k$) and diagonal scale $S_k$.
Rendering uses differentiable rasterization, where Gaussians are alpha-blended along each ray:
\[
C=\sum_{d=1}^{D} c_d\,\alpha_d \!\!\prod_{j<d}(1-\alpha_j),
\]
with $c_d$ evaluated from SH coefficients and $\alpha_d$ incorporating opacity and projected density. 

\subsection{Template 3DGS Face Model Construction}
\label{sec:canonical}
Directly inferring the parameters of a full 3DGS model from a single image is highly under-constrained due to the large number of parameters involved: each Gaussian has 59 parameters including geometry (center, scale, rotation), opacity, and 48 SH appearance coefficients. 
Inspired by classical morphable models~\cite{blanz2023morphable}, we reduce this complexity by predicting only the \emph{residual deformations} required to adapt a data-driven template 3DGS model $\mathcal{T}$ to the input image.

Following GaussianAvatars~\cite{qian2024gaussianavatars}, we place one Gaussian at the center of each triangular face of the FLAME mesh~\cite{FLAME:SiggraphAsia2017} and add additional Gaussians for the upper and lower teeth. 
This produces a canonical set of $K$ mesh-attached Gaussians that share consistent semantic meaning across subjects and provide a stable geometric reference for learning. 
This design also enables efficient real-time animation, as described in Section~\ref{sec:animation}.

As shown in Fig.\ref{fig:pipeline}(a), to construct the template $\mathcal{T}$, we first optimize individual 3DGS models $\{\mathcal{M}_i\}_{i=1}^N$ for all training subjects, each initialized with Gaussians placed at these consistent FLAME-aligned locations. 
We then apply standard 3DGS optimization~\cite{ye2025gsplat} to recover subject-specific geometry and appearance. 
Finally, the template $\mathcal{T}$ is obtained by averaging the parameters of corresponding Gaussians across subjects. 
Compared to random initialization or joint template optimization~\cite{saunders2025gasp,qian2024gaussianavatars,Yan_2025_WACV,liu2024dynamic}, this simple averaging procedure yields a compact, stable prior requiring only small subject-specific residuals during reconstruction. We show the effectiveness of the average template in Sec. \ref{sec:ablation}, and a more detailed analysis in Supplementary.

\subsection{Encoder-Decoder Network}

The encoder–decoder network maps an input image $I$ to residual Gaussian parameters that deform the template $\mathcal{T}$ into a subject-specific 3DGS model. 
Because $I$ can appear under arbitrary viewpoints, the encoder is designed to produce a \emph{pose-invariant} latent code that captures only identity-specific information.

Training the encoder and decoder end-to-end leads to a degenerate solution where the network collapses to predicting an average face. 
To avoid this, we adopt a two-stage training strategy. 
We first train the decoder while treating the latent codes for all training subjects as learnable variables. 
This stage forces the decoder to learn a smooth and generalizable latent space for predicting Gaussian residuals, rather than memorizing per-identity solutions.

After the decoder is trained, we freeze it and train the encoder to map images into the learned latent space. 
The encoder is guided by features from a large-scale face recognition model to ensure pose invariance and strong identity discrimination. 
This staged design yields an encoder that generalizes reliably to unseen identities and expressions, while the decoder provides stable geometry prediction anchored to the canonical template. 

\subsubsection{Decoder Design and Optimization}

The decoder $h_\theta$ maps a pose-invariant identity code $w\!\in\!\mathbb{R}^{|w|}$ and a Gaussian embedding 
$e_k\!\in\!\mathbb{R}^{|e|}$ to residual parameters 
$\Delta G_k=\{\Delta\mu_k,\Delta s_k,\Delta q_k,\Delta\alpha_k,\Delta c_k\}$ for Gaussian $k$. 
The identity code $w$ captures subject-dependent properties, while $e_k$ provides localized context for each Gaussian.

For appearance parameters (scale, rotation, opacity, SH coefficients), residuals are applied to the template values. 
For positions, we add $\Delta\mu_k$ to the FLAME-tracked canonical location $\bar{\mu}_k$.
Thus the final parameters are  
$\mu_k=\bar{\mu}_k+\Delta\mu_k$,  
$s_k=s_k^\mathcal{T}\!\cdot\!\exp(\Delta s_k)$,  
$q_k=\text{normalize}(q_k^\mathcal{T}+\Delta q_k)$,  
$\alpha_k=\sigma(\text{logit}(\alpha_k^\mathcal{T})+\Delta\alpha_k)$,  
and  
$c_k=c_k^\mathcal{T}+\Delta c_k$.

We implement the decoder with a shallow MLP. 
Gaussian embeddings $\{e_k\}$ are initialized using sinusoidal positional encodings of their canonical FLAME coordinates, and identity codes $\{w_i\}_{i=1}^N$ are initialized from a standard Normal distribution. 
During decoder pretraining, we jointly optimize $h_\theta$, $\{w_i\}$, and $\{e_k\}$ using
\begin{align}
\mathcal{L}_{\text{dec}}
&= \lambda_1 \mathcal{L}_{\text{LPIPS}}
+ \lambda_2 \mathcal{L}_1
+ \lambda_3 \mathcal{L}_{\text{SSIM}} \nonumber\\
&\quad + \lambda_4 \mathcal{L}_2(\Delta\mu)
+ \lambda_5 \mathcal{L}_2(\Delta s),
\end{align}
computed between rendered predictions and ground-truth images. 
LPIPS, SSIM, and $L_1$ promote perceptual and photometric accuracy, while the $L_2$ regularizers keep residuals small and stable.

\subsubsection{Encoder Design and Optimization}
\label{sec:encoder}
The encoder $g_\phi$ predicts a pose-invariant latent code for the face depicted in $I$, from which the decoder can infer a full 3DGS representation. 
We construct it as a composition $g_\phi = g_{\text{MLP}} \circ g_{\text{FR}}$, where $g_{\text{FR}}$ is a pretrained face recognition backbone and $g_{\text{MLP}}$ is a lightweight projection network. 
The face recognition model provides identity features that are invariant to viewpoint, while the projection MLP maps these features to the latent space used by the decoder.

\begin{figure*}[!t]
    \includegraphics[width=\textwidth]{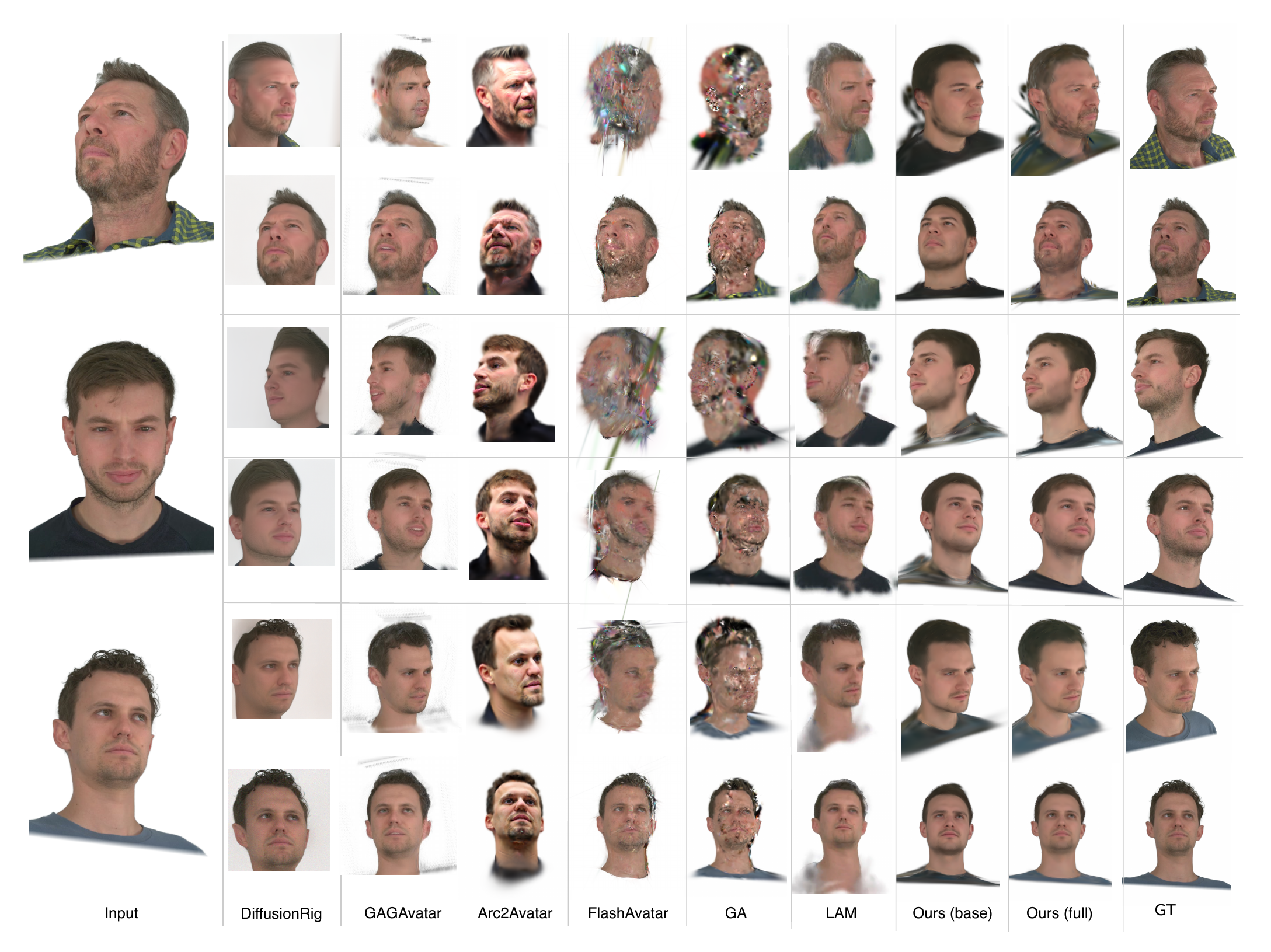}
   \caption{\textbf{Qualitative comparison on single-image novel-view synthesis.} 
Given a single arbitrary-view input (left), we compare \alg{} (base and full) with DiffusionRig~\cite{ding2023diffusionrig}, 
GAGAvatar~\cite{chu2024generalizable}, LAM~\cite{he2025lam}, Arc2Avatar~\cite{gerogiannis2025arc2avatar}, 
FlashAvatar~\cite{xiang2024flashavatar}, and GaussianAvatars (GA)~\cite{qian2024gaussianavatars}. 
GAGAvatar and Arc2Avatar operate in their own canonical spaces; following prior work, we align their outputs 
to our coordinate frame via PnP (details in Supplementary), though small residual shifts may remain. 
Diffusion-based and feed-forward baselines struggle under large input poses, often producing blurry textures, synthetic-looking faces, or distorted geometry. 
GA and FlashAvatar, which require multi-view fitting, degrade noticeably when extended to the single-view setting. 
In contrast, \alg{} maintains coherent geometry and identity across wide viewpoint changes; the full model further sharpens appearance through a lightweight $3$-second refinement stage. 
Additional examples, including more poses, expressions, and identity-similarity metrics, are provided in the Supplementary.}
    \label{fig:comparison}
\end{figure*}

Given the precomputed identity codes $\{w_i\}_{i=1}^N$ learned during decoder training, we train $g_\phi$ to map all images of the same subject to their code. The encoder loss is
\begin{align}
\mathcal{L}_{\text{enc}}(w, \hat{w}) 
= \mathcal{L}_{2}(w, \hat{w})
+ \lambda_{\text{cos}}\, \mathcal{L}_{\text{cos}}(w, \hat{w}),
\end{align}
where $\hat{w}=g_\phi(I)$ is the predicted embedding and $\mathcal{L}_{\text{cos}}$ is the cosine distance. 
This training strategy encourages the encoder to generalize to unseen identities and arbitrary poses.

\subsection{Appearance Refinement at Inference Time}

Given an input image $I$ from an arbitrary viewpoint at inference time, a single encoder–decoder pass first predicts residuals that deform the canonical template, producing a stable and 3D-consistent geometry anchored by the learned prior. We then apply a lightweight refinement step that optimizes only the latent code $w$ and the decoder’s appearance outputs for $\sim 300$ iterations ($\sim 3$ seconds on an A100). Crucially, unlike methods that directly optimize Gaussian parameters from a single view~\cite{saunders2025gasp,he2025lam} -- which can distort geometry and harm multi-view consistency -- this refinement does not update Gaussian parameters explicitly. Instead, all adjustments flow through the encoder–decoder prior, ensuring that any changes remain consistent with the learned canonical geometry and do not introduce view-specific artifacts. Once reconstruction is complete, the resulting Gaussians can be rendered in real time using standard 3DGS rasterization.


\subsection{Animation and Reenactment}
\label{sec:animation}
Our animation module follows the mesh-attached Gaussian design introduced in the template construction stage (Sec.~\ref{sec:canonical}), similar to GaussianAvatars~\cite{qian2024gaussianavatars}. 
Each Gaussian is anchored to a FLAME face-center location (plus teeth anchors) and inherits fixed skinning weights $\{w_{k,j}\}_{j=1}^{J}$ from the FLAME mesh.

Given target FLAME expression or pose parameters, we obtain the joint transformations $\{A_j\}_{j=1}^{J}$ and animate each Gaussian using standard linear blend skinning (LBS). 
For Gaussian $k$, the local canonical parameters $(\mu'_k,\sigma'_k,\mathbf{r}'_k)$ are mapped to global space via
\[
(\mu_k^{\mathrm{anim}},\sigma_k^{\mathrm{anim}},\mathbf{r}_k^{\mathrm{anim}})
= 
\mathcal{T}_{\mathrm{local}\rightarrow\mathrm{global}}
\bigl((\mu'_k,\sigma'_k,\mathbf{r}'_k),\, T_k \bigr),
\]
where $\mathcal{T}_{\mathrm{local}\rightarrow\mathrm{global}}$ denotes the standard SE(3)-based Gaussian frame transform, and $T_k = \sum_{j=1}^{J} w_{k,j} A_j$. This updates position, rotation, and scale consistently with the FLAME-driven deformation. 

This procedure yields smooth expression and pose changes without additional optimization, enabling real-time reenactment once the 3DGS model is reconstructed.

\section{Experiments \& Results}
We evaluated \alg{} on single-view 3D face reconstruction using the Nersemble dataset~\cite{kirschstein2023nersemble}, which contains 417 identities captured simultaneously from 16 calibrated viewpoints spanning frontal to extreme profile poses. Each identity is recorded under multiple expressions. We used 410 identities for training and held out 7 identities for testing. For training, we sampled 6 expressions per identity and assigned each identity–expression pair a distinct latent code.

\textbf{Baselines.}  
We compare \alg{} (base and full) against recent state-of-the-art 3DGS and neural avatar methods:  
(1) \textit{Optimization-driven}: GaussianAvatars~\cite{qian2024gaussianavatars}, FlashAvatar~\cite{xiang2024flashavatar};  
(2) \textit{Feed-forward}: GAGAvatar~\cite{chu2024generalizable}, LAM~\cite{he2025lam};  
(3) \textit{Diffusion-based}: Arc2Avatar~\cite{gerogiannis2025arc2avatar}, DiffusionRig~\cite{ding2023diffusionrig}.  
We use the official implementations and recommended configurations for all methods. Because Arc2Avatar, GAGAvatar, and LAM operate in their own canonical spaces, we estimate camera intrinsics and extrinsics by PnP-aligning their reconstructed geometry to our canonical template (see Supplementary). Throughout the experiments, Ours(base) denotes the feed-forward encoder–decoder prediction (geometry only), and Ours(full) includes the subsequent appearance refinement.

\textbf{Metrics.}  
We evaluate reconstruction accuracy using PSNR, SSIM~\cite{wang2004image}, LPIPS~\cite{zhang2018unreasonable}, and Identity Similarity~\cite{deng2018arcface}. For each test identity, we use one of the 16 views as the input and evaluate rendering quality on the remaining 15 views, repeating this for all input viewpoints.  
All runtime numbers are measured on a single NVIDIA A100 (40GB).

\textbf{Implementation Details.}  
We constructed the template $\mathcal{T}$ from the 410 subject-specific 3DGS models optimized for 7{,}000 iterations using all 16 views, each trained with one randomly sampled expression. We used VHAP~\cite{qian2024vhap} to extract FLAME parameters and camera poses. Following GaussianAvatars, we place one Gaussian at the center of each FLAME mesh face (9{,}976 faces) and add 168 Gaussians for the upper and lower teeth, resulting in $K = 10{,}144$ mesh-attached Gaussians with consistent semantic correspondence across all subjects. Please refer to Supplementary for additional implementation and training details.

\subsection{Novel View Reconstruction Results}
We first present qualitative comparisons in Fig.~\ref{fig:comparison} for three sample test cases. GAGAvatar, LAM, and DiffusionRig exhibit strong degradation under non-frontal inputs, producing broken geometry, scattered points, or blurry outputs. Arc2Avatar often yields synthetic facial textures and may alter expressions (e.g., adding an open mouth). FlashAvatar and GaussianAvatars (GA), which rely on multi-view optimization, struggle in the single-view setting and produce noisy or incomplete reconstructions for challenging viewpoints. These two methods can be viewed as refinement-only analogs of our system: they optimize Gaussian parameters directly without a geometric prior, highlighting the role of \alg{}’s encoder–decoder stage in establishing stable, 3D-consistent structure from a single image.

In contrast, \alg{} delivers stable geometry and consistent identity across all novel views. The feed-forward prediction (Ours(base)) already provides coherent 3D structure from a single input, while the refinement stage (Ours(full)) further enhances appearance, capturing subtle facial details, hair, and clothing with improved fidelity. Together, these stages achieve high-quality novel-view synthesis that remains accurate across large viewpoint changes.
\begin{table}[t!]
\centering
\small 
\caption{\textbf{Quantitative comparison on novel-view reconstruction.} 
\alg{} achieves the best performance across all metrics. 
Methods marked with an asterisk (*) may have slightly underestimated scores due to minor residual misalignment after PnP-based canonical alignment.}
\label{tab:comparison}
\begin{tabular}{l|c@{\hspace{1em}}c@{\hspace{1em}}c@{\hspace{1em}}c}
\hline
\textbf{Method} & \textbf{PSNR} $\uparrow$ & \textbf{SSIM} $\uparrow$ & \textbf{LPIPS} $\downarrow$  & \textbf{ID Sim.}
$\uparrow$\\
\hline
DiffusionRig   & 14.21 & 0.70 & 0.29 & 0.65\\
GAGAvatar$^*$       & 15.83 & 0.73 & 0.33 & 0.76 \\
Arc2Avatar$^*$       & 14.48 & 0.78 & 0.30 & 0.61 \\
FlashAvatar     & 13.99 & 0.76 & 0.32 & 0.35 \\
GaussianAvatars & 16.39 & 0.79 & 0.30 & 0.39\\
LAM & 14.13 & 0.81 & 0.34 & 0.77\\
\hline
Ours (base) & 21.17 & 0.89 & 0.22 & 0.70\\
Ours (full)        & \textbf{24.01} & \textbf{0.91} & \textbf{0.19} &  \textbf{0.81}\\
\hline
\end{tabular}
\end{table}

Table~\ref{tab:comparison} summarizes quantitative results for novel-view reconstruction. Ours (full) achieves the best performance across all metrics (24.01,dB PSNR, 0.91 SSIM, 0.19 LPIPS and 0.81 ID similarity), substantially outperforming all baselines. Ours (base) -- the feed-forward encoder–decoder prediction -- already provides strong geometry (21.17 dB PSNR, 0.89 SSIM) before the refinement phase. 

\begin{figure}[t!]
    \includegraphics[width=0.48\textwidth]{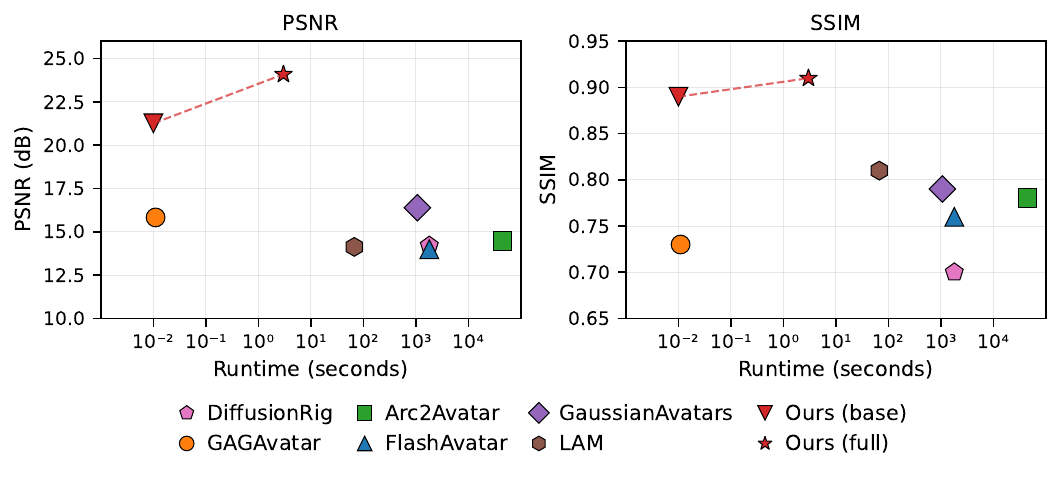}
    \caption{\textbf{Reconstruction quality vs.\ runtime.} 
    Ours (base) produces strong feed-forward reconstructions (21.17\,dB PSNR, 0.89 SSIM), while Ours (full) achieves state-of-the-art quality (24.01\,dB, 0.91 SSIM) with only $\sim$3 seconds of refinement.}
    \label{fig:quant}
\end{figure}

\subsection{Runtime Results}
Fig.~\ref{fig:quant} presents the trade-off between reconstruction quality and fitting time for different methods. GAGAvatar and LAM operate in (milli-)seconds but produce far lower accuracy. Diffusion-based (DiffusionRig, Arc2Avatar) and optimization-driven (FlashAvatar, GaussianAvatars) methods require seconds-to-hours of per-subject optimization while still trailing our method in accuracy. Ours (full) achieves the best overall reconstruction with only $\sim$3 seconds of refinement, representing a substantially better quality–speed trade-off than all prior approaches. Ours (base) runs in a single feed-forward pass and already reaches strong accuracy, highlighting the efficiency and stability of the encoder–decoder geometry prior.

\subsection{Self- and Cross-Reenactment}
\begin{figure}[t!]
    \includegraphics[width=0.48\textwidth, trim=0 10 0 0, clip]{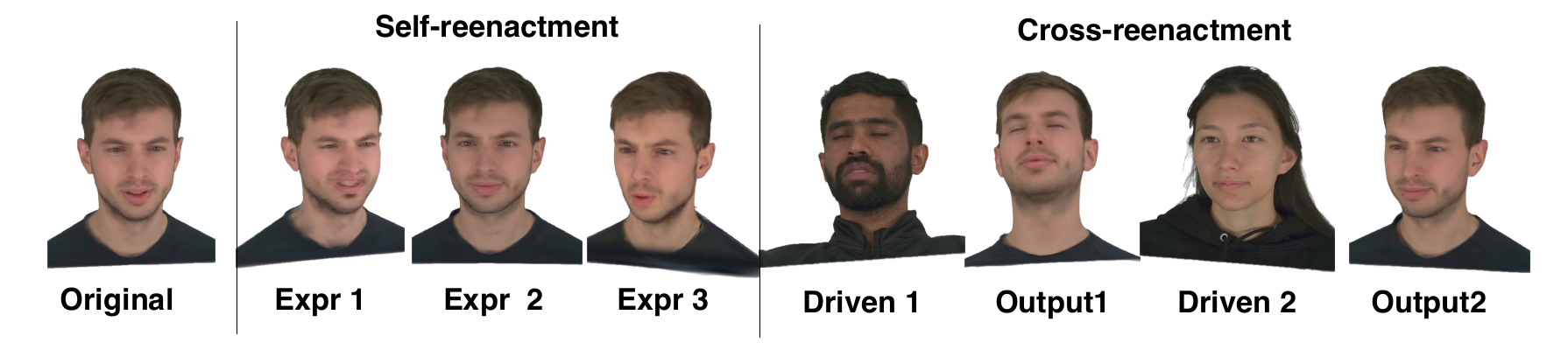}
    \caption{\textbf{Self- and cross-reenactment.} 
    Starting from a single reconstructed face (left), \alg{} can reproduce expressions from the same subject (self) or transfer expressions from another subject (cross) by driving the Gaussians with FLAME parameters. 
    Identity remains stable while expressions are well reproduced.}
    \label{fig:reenactment}
\end{figure}

Fig.~\ref{fig:reenactment} shows both self- and cross-reenactment results. 
In the self-reenactment examples, we drive the reconstructed subject using FLAME expression parameters extracted from other frames of the same identity. 
The outputs follow the target expressions while maintaining the subject’s overall geometry and appearance. 
In the cross-reenactment examples, we use FLAME parameters from a different “driver’’ subject. 
The transferred expressions are reproduced on the source face without altering its identity, and the deformations remain stable across different expressions. 
These examples illustrate that the FLAME-guided deformation model allows \alg{} to reproduce a range of expressions from a single reconstructed identity.

\subsection{Out-Of-Distribution Identities}
\begin{figure}[t!]
    \includegraphics[width=0.48\textwidth, trim=0 25 0 10, clip]{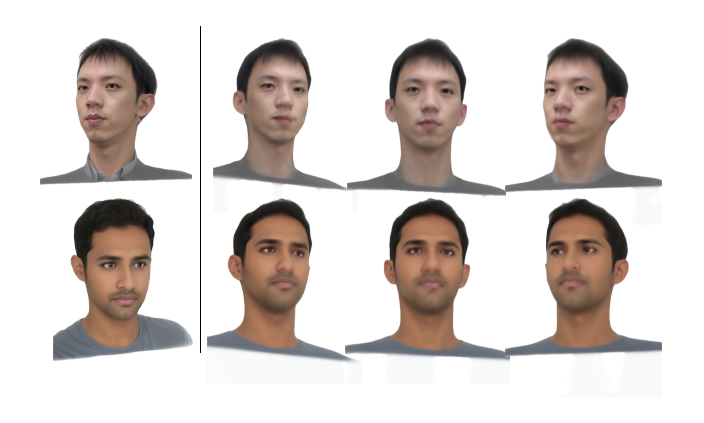}
    \caption{\textbf{Generalization to out-of-distribution identities.} 
    Left column = single input image. The remaining columns = novel-view renderings from the reconstructed 3DGS model.  
    \alg{} produces stable, identity-preserving reconstructions for unseen subjects and maintains consistent geometry across views.}
    \label{fig:ood}
\end{figure}

We further tested \alg{} on identities outside of the Nersemble dataset. We obtained the FLAME parameters for these images using EMOCA~\cite{danvevcek2022emoca}, ensuring consistent alignment with our canonical template. As shown in Fig.~\ref{fig:ood}, \alg{} reconstructs clean, identity-preserving 3DGS models from a single image of OOD identities. We provide additional OOD subjects and viewpoints in Supplementary.

\subsection{Latent Space Structure and Decoder Analysis}
\label{sec:traversal}
\begin{figure*}[!t]
    \includegraphics[width=\textwidth]{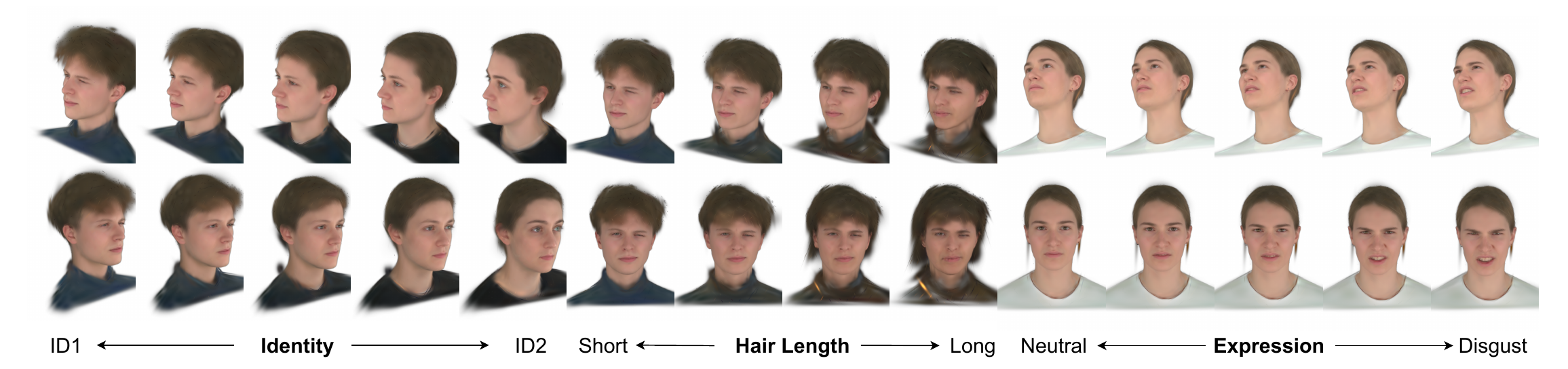}
    \caption{\textbf{Latent space interpolation and attribute traversals.}
    \textbf{Left:} Identity interpolation between two codes produces smooth, realistic transitions in geometry and appearance. 
    \textbf{Middle and Right:} Moving along learned attribute directions (e.g., hair length, expression intensity) yields consistent edits across views while preserving identity, illustrating that the decoder has learned a meaningful latent space.}
    \label{fig:traversal}
\end{figure*}

We analyzed the latent space learned by the decoder to verify that it captures a smooth and generalizable representation rather than memorizing training identities. Fig.~\ref{fig:traversal}(left)) presents identity interpolation traversals between two latent codes. The intermediate reconstructions vary smoothly in geometry and appearance and remain consistent across novel viewpoints, indicating a smooth latent space. We also explored global attribute directions in the latent space (e.g., hair length, expression intensity). Traversing a code along these directions produces coherent edits while preserving identity and multi-view consistency (Fig.~\ref{fig:traversal}(middle, right)). We provide the procedure for estimating these directions and more results in Supplementary.

\begin{table}[t!]
\centering
\small
\setlength{\tabcolsep}{6pt}
\caption{\textbf{Ablation results on \alg{} design choices.} All results are reported for full version unless noted.}
\begin{tabular}{lccc}
\hline
\textbf{Setting} & \textbf{PSNR} $\uparrow$ & \textbf{SSIM} $\uparrow$ & \textbf{Runtime (ms)} \\
\hline
\multicolumn{4}{l}{\textit{Number of Gaussians $K$} (\textbf{Default:} $10{,}144$)} \\
5023           & 20.94 & 0.84 & 9 \\
40{,}000       & 25.58 & 0.93 & 29 \\
\hline
\multicolumn{4}{l}{\textit{Loss Weights} (\textbf{Default:} Eq.~(1)--(2))}\\
w/o SSIM (dec.)       & 22.03 & 0.88 & 10 \\
w/o CosSim (enc.)     & 22.27 & 0.89 & 10 \\
\hline
\multicolumn{4}{l}{\textit{Gaussian Init.} (\textbf{Default:} Average Template)}\\
Random Init           & 21.92 & 0.88 & 10 \\
Joint-Opt.            & 22.84 & 0.89 & 10 \\
\hline
\multicolumn{4}{l}{\textit{Refinement Iterations} (\textbf{Default:} $300$)}\\
0 (feed-forward)      & 21.17 & 0.89 & 10 \\
600                   & 24.32 & 0.91 & 6000 \\
\hline
\textbf{Ours (full)}  & 24.01 & 0.91 & 3000 \\
\hline
\end{tabular}
\label{tab:ablation}
\end{table}

\subsection{Ablation Studies}
\label{sec:ablation}
We finally present an ablation study of \alg{}'s key design choices in Table~\ref{tab:ablation}. We first find that fewer Gaussians ($K{=}5023$) lose detail, while a larger number ($K{=}40{,}000$) improves fidelity but slows inference. Our default $K{=}10144$ strikes a balance. Second, removing SSIM (decoder) or cosine distance (encoder) consistently degrades final rendering quality. Third, initialization with the learned template model yields the most stable and accurate reconstructions, outperforming joint-optimizing a canonical template during decoder training or random initialization. Fourth, test-time refinement improves results, though long runs (e.g., 600 steps) give diminishing returns relative to the temporal cost. Full results, including visual comparisons, are in Supplementary.

\section{Discussion and Conclusion}
\alg{} provides a practical solution for single-image 3D face reconstruction across large pose variations. 
The two-stage pipeline: an encoder–decoder feed-forward prediction followed by a lightweight refinement, allows us to recover clean geometry in one pass and improve appearance with only \(\sim\!3\) seconds of additional computation. 
First, a key component of this design is the mesh-attached average Gaussian template, which establishes stable semantic correspondences across subjects and makes residual prediction well conditioned. 
Our ablation studies confirm that this canonical parameterization is critical for reliable geometry estimation and avoids the instability. The encoder–decoder architecture further contributes to generalization. 
By encouraging a pose-invariant latent representation during training, the model can handle a wide variety of inputs and maintain identity consistency for unseen subjects. 
Together, these components enable \alg{} to produce high-quality reconstructions while keeping the overall system simple and efficient.

\textbf{Limitations.}  
\alg{} inherits structural constraints from FLAME and the face-recognition backbone used by the encoder. FLAME does not model long hair, fine strands, or clothing, and most face-recognition models crop tightly around the face. As a result, subjects with long hairstyles, prominent bangs, hats, or high-variation clothing, especially common in female subjects, may exhibit smoothed or incomplete reconstructions around these regions, we share a failure case study in the Supplementary. These limitations are shared by most current 3DGS-based avatar systems and point toward the need for more expressive priors and broader training data. 
In addition, the Nersemble dataset itself has limited demographic and appearance diversity, which constrains the range of identities and hairstyles that current methods can reliably model. Richer multi-view datasets would help explore the full capabilities and limitations of single-image 3DGS reconstruction in this space.


\clearpage
{
    \small
    \bibliographystyle{ieeenat_fullname}
    \bibliography{main}
}
\end{document}